\documentclass{article} % For LaTeX2e
\usepackage{iclr2016_conference,times}
\usepackage{url}
\usepackage{graphicx}
\usepackage{amssymb}
\usepackage{amsthm}
\usepackage{subfigure}
\usepackage{tikz}
\usepackage{lipsum}

\title{Structured Sparse Convolutional Autoencoder }

\author{Ehsan~Hosseini-Asl
\\
Department of Electrical and Computer Engineering\\
University of Louisville\\
Louisville, KY 40208, USA \\
\texttt{\{ehsan.hosseiniasl\}@gmail.com} \\
}

\begin{document}

\maketitle

\begin{abstract}
This paper aims to improve the feature learning in Convolutional Networks (Convnet) by capturing the structure of objects. A new sparsity function is imposed on the extracted featuremap to capture the structure and shape of the learned object, extracting interpretable features to improve the prediction performance. The proposed algorithm is based on organizing the activation within and across featuremap by constraining the node activities through $\ell_{2}$ and $\ell_{1}$ normalization in a structured form.
\end{abstract}

\section{Introduction}
Convolutional net (Convnet)~\cite{lecun1998gradient} have shown to be powerful models in extracting rich features from high-dimensional images. They employ hierarchical layers of combined convolution and pooling to extract compressed features that capture the intra-class variations between images. The purpose of applying pooling over neighbor activations in featuremaps of Convnet is to break the spatial correlation of neighboring pixels, and to improve the scale and translation invariant features learned by Convnet. This also helps in learning filters for generic feature extraction of low-mid-high level of concepts, such as edge detectors, geometric shapes, and object class ~\cite{krizhevsky2012imagenet,donahue2013decaf,zeiler2010deconvolutional,zeiler2014visualizing}.

Several regularization techniques have been proposed to improve feature extraction in Counvnet and to overcome overfitting in large deep networks with many parameters. A dropout technique in~\cite{srivastava2014dropout} is based on randomly dropping hidden units with its connnection during training to avoid co-adaptaion or redundant filter training. This method resemble averaging over ensemble of sub-models, where each sub-model is trained based on a subset of parameters. A maxout neuron is proposed in~\cite{goodfellow2013maxout} while a maxout neuron, with the maximum of activity across featuremaps is computed in Counvnets. Maxout networks have shown to improve the classification performance by building a convex an unbounded activation function, which prevents learning dead filters. A winner-take-all method is employed in~\cite{makhzani2014winner} to reduce or eliminate redundant and delta type filters in pretraining of Counvnet using Convolutional AutoEncoder (CAE), by taking the maximum activity inside featuremap in each training step. 

%Patch-based unsupervised learning is proposed for pretraining of Counvnet, where randomly selected patches from unlabeled dataset are used to extract global patterns and features using training an Autoencoder. Then the trained receptive fields are used to pretrain Convnets.

Sparse feature learning is a common method for compressed feature extraction in shallow encoder-decoder-based networks, i.e. in sparse coding~\cite{hoyer2002non,hoyer2004non,olshausen1996emergence,olshausen1997sparse}, in Autoencoders (AE)~\cite{ng2011ufldl}, and in Restricted Boltzmann Machines (RBM)~\cite{poultney2006efficient,marc2007sparse}. Bach et al.~\cite{bach2012structured} organize $\ell_{1}$ sparsity in a structured form to capture interpretable features and improve prediction performance of the model. In this paper, we present a novel Structured Model of sparse feature extraction in CAE that improves the performance of feature extraction by regularizing the distribution of activities inside and across featuremaps. We employ the idea of sparse filtering~\cite{ngiam2011sparse} to regularize the activity across featuremaps and to improve sparsity within and across featuremaps. The proposed model is using $\ell_{2}$ and $\ell_{1}$ normalization on the featuremap activations to implement part-based feature extraction.

\section{Model}
\label{sec:model}
In this section, the model of Structured Sparse CAE (SSCAE) is described. CAE consists of convolution/pooling/nonlinearity based encoding and decoding layers, where the feature vector is represented as featuremaps, i.e. hidden output of the encoding layer.  

\begin{figure}[htb!]
	\begin{minipage}{0.5\linewidth}
		\centering
		\includegraphics[width=7cm]{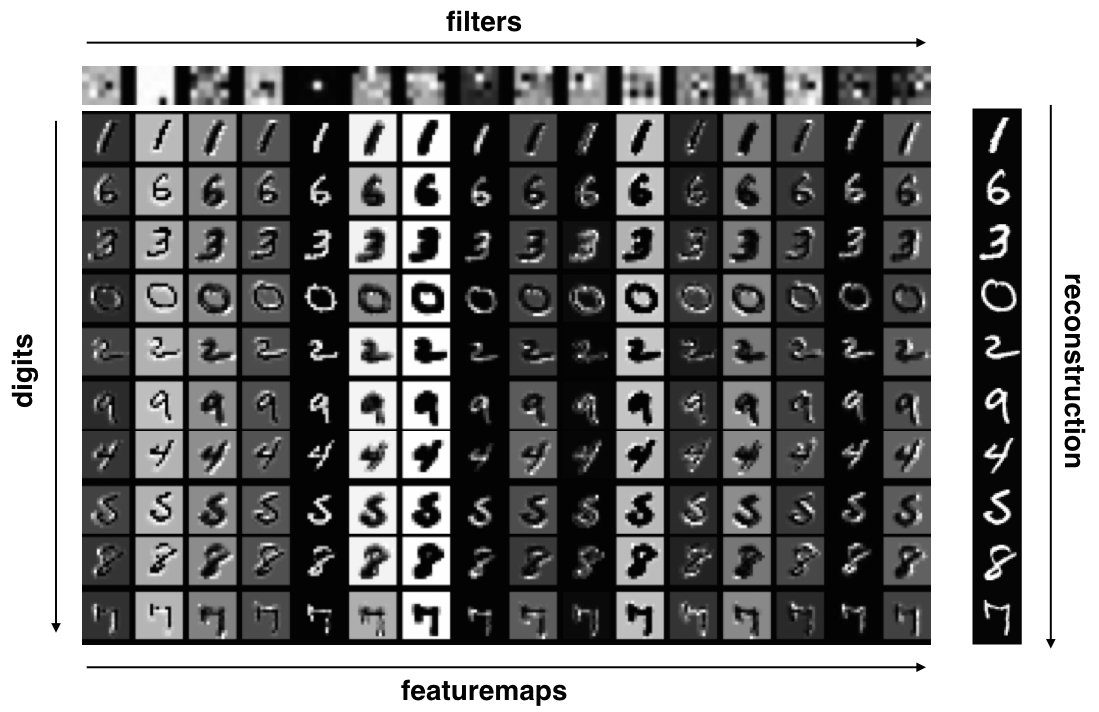}
		
		{{\footnotesize (a) CAE}}
	\end{minipage}
	\begin{minipage}{0.5\linewidth}
		\centering
		\includegraphics[width=7cm]{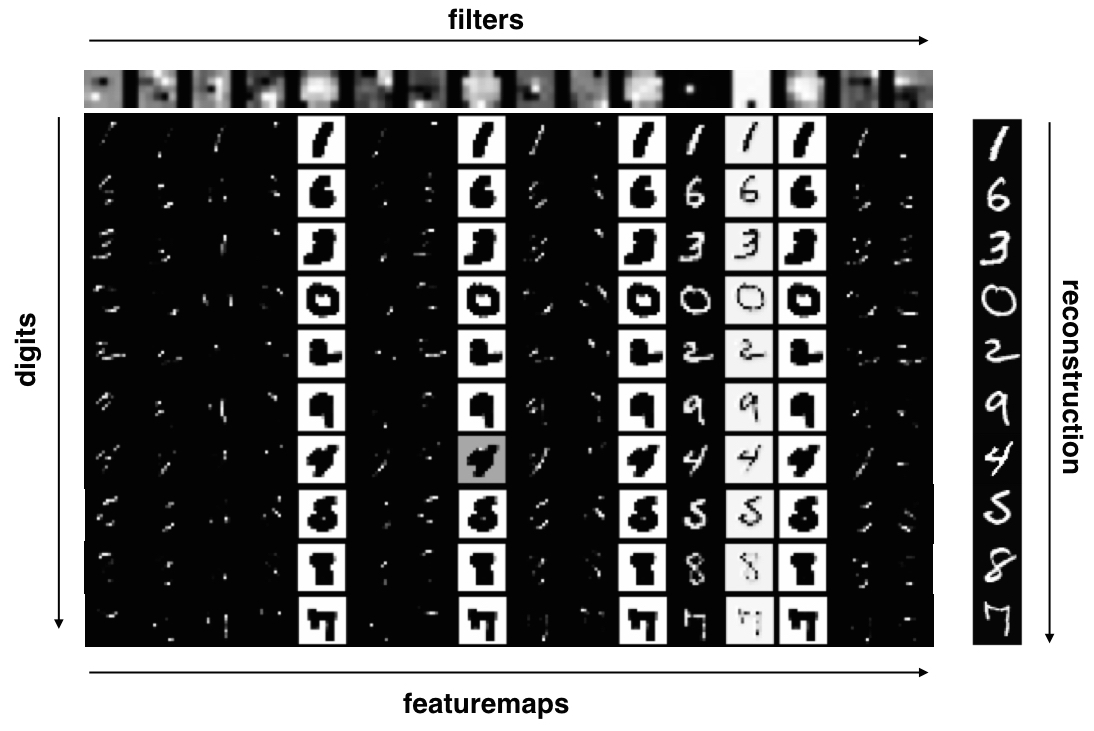}
		
		{{\footnotesize (b) SSCAE}}
	\end{minipage}
	\caption{16 example filters ($\textbf{W}^{k\in [1,\ldots,16]}=[w_{ij}]_{5\times 5}$) and featuremaps ($\textbf{h}^{k\in [1,\ldots,16]}=[h^{k}_{ij}]_{24\times 24}$), with feature vectors ($\textbf{h}_{ij}=[h_{ij}^{k}]_{1\times 16}$), extracted from non-whitened MNIST with sigmoid nonlinearity and no pooling using (a) CAE, (b) SSCAE. Effect of sparse feature extraction using SSCAE is shown w/o pooling layer. Digits are input pixelmaps $28\times 28$, $n=16$ for this example.}
	\label{fig:basic}
\end{figure}

\begin{figure*}[htb!]
	%	\vspace{-10mm}
	\centering
	\includegraphics[width=12cm]{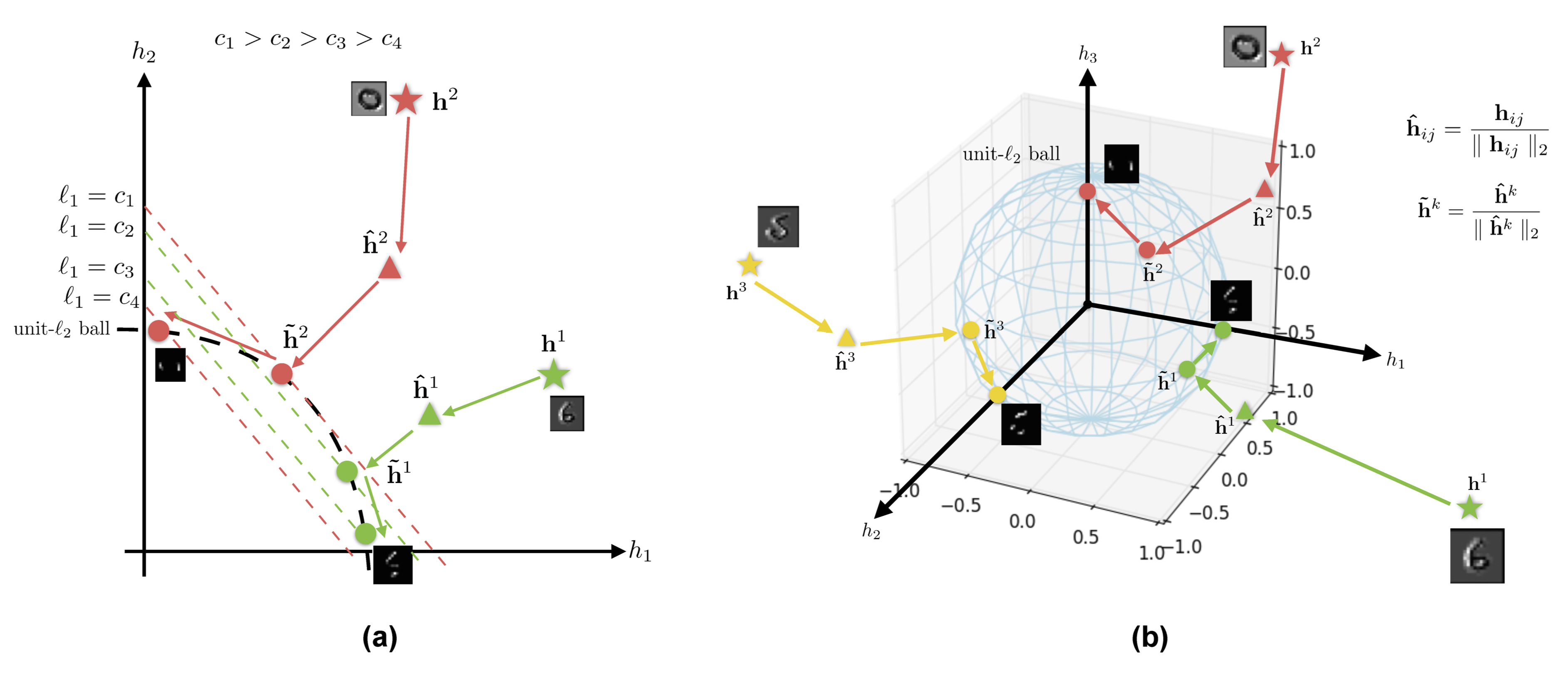}
	%	\vspace{10mm}
	\caption{Structured Sparsity on illustration on (a) two-dimensional and (b) three-dimensional space for featuremaps ($\mathbf{h}^{1}$, $\mathbf{h}^{2}$, $\mathbf{h}^{3}$) of MNIST dataset. Each example is first projected onto the unit $\ell_{2}$-ball and then optimized for $\ell_{1}$ sparsity. The unit $\ell_{2}$-ball is
		shown together with level sets of the $\ell_{1}$-norm. Notice that the sparseness of the features (in the $\ell_{1}$ sense) is maximized when the examples are on the axes~\cite{ngiam2011sparse}.}
	\label{fig:sparse}
\end{figure*}

In CAE with $n$ encoding filters, the featuremaps $\mathbf{h}^{k\in n}$ are computed based on a convolution/pooling/nonlinearity layer, with nonlinear function applied on the pooled activation of convolution layer, as in Eq.~\ref{eq:hidden}. 

\begin{equation}
		\mathbf{h}^{k\in n}=f(x\ast\mathbf{W}^{k\in n}+b^{k\in n})
		\label{eq:hidden}
\end{equation}
where $\mathbf{W}^{k\in n}$ and $b^{k\in n}$ are the filter and bias of $k$-th featuremap, respectively. We refer to $h_{ij}^{k}$ as single neuron activity in $k$-th featuremap $\mathbf{h}^{k}$, whereas $\mathbf{h}_{ij}=\left[h_{ij}^{k}\right]^{k\in n}$ is defined as a feature vector across featuremaps $\mathbf{h}^{k\in n}$ as illustrated in Fig.~\ref{fig:basic}.

In SSCAE, the featuremaps $\mathbf{h}^{k\in n}$ are reqularized and sparsified to represent three properties; (\emph{i}) \textit{Sparse feature vector $\mathbf{h}_{ij}$};
 (\emph{ii}) \textit{Sparse neuronal activity $h_{ij}^{k}$ within each of the $k$-th featuremap $\mathbf{h}^{k}$};  (\emph{iii})\textit{Uniform distribution of feature vectors $\mathbf{h}_{ij}$}. 

In (\emph{i}), sparsity is imposed on feature vector $\mathbf{h}_{ij}$ to increase diversity of features represented by each featuremap, i.e. each $\mathbf{h}^{k\in n}$ should represent a distinguished and discriminative characteristic of the input, such as different parts, edges, etc. This property is exemplified in Fig.~\ref{fig:basic}(b) with digits decomposed into parts across featuremaps $\mathbf{h}^{k\in n}$. As stipulated in (\emph{ii}), sparsity is imposed on each featuremap $\mathbf{h}^{k\in n}$ to only contain few non-zero activities $h^{k}_{ij}$. This property is encouraged for each featuremap to represent a localized feature of the input. Fig.~\ref{fig:basic}(b) shows property (\emph{i}) for MNIST dataset, where each featuremap is a localized feature of a digit, wherein Fig.~\ref{fig:basic}(a) shows extracted digit shape-resemblance featurs, a much less successful and non-sparse outcome compared to Fig.~\ref{fig:basic}(b). Fig.~\ref{fig:sparse} also depicts the technique for numerical sparsification of each featuremap. The property (\emph{iii}) is imposed on activation features $\mathbf{h}_{ij}$ to have similar statistics with uniform activity. In other words, $\mathbf{h}_{ij}$ will be of nearly equal or uniform activity level, if they lie in the object spatial region, or non-active, if not. Uniform activity also improves the generic and part-based feature extraction where the contributing activation features $\mathbf{h}_{ij}$ of digits, i.e. $\mathbf{h}_{ij}$, fall within convolutional region of digits and filters $\mathbf{W}^{k\in n}$ show uniform activity level, which results in generic and part-based features. 

\begin{figure}[htb!]
	\centering
	\includegraphics[width=1\linewidth]{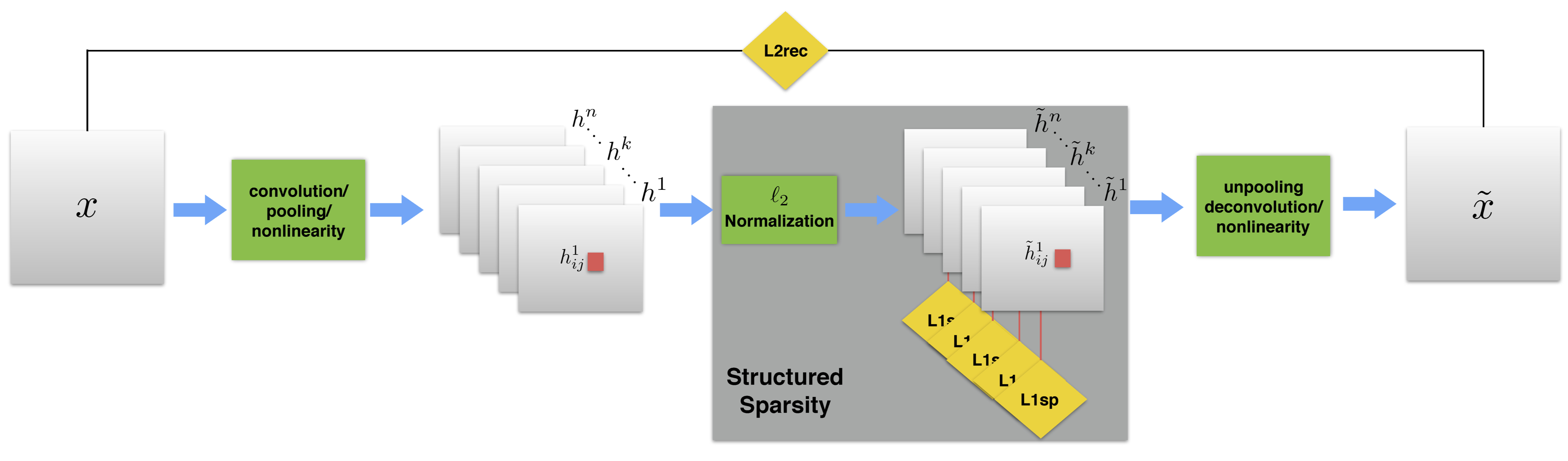}
	\caption{Model architecture of Structured Sparse Convolutional AutoEncoder (SSCAE)}
	\label{fig:model}
\end{figure}

To enforce the aformentioned sparsity properties in CAE models, we have used the combination of $\ell_{2}$ and $\ell_{1}$ normalization on $\mathbf{h}^{k\in n}$ of Eq.~\ref{eq:hidden}, as proposed in~\cite{ngiam2011sparse}, and as shown in Fig.~\ref{fig:model}. In SSCAE, a normalization layer is added on the encoding layer, where the normalized featuremaps $\tilde{\mathbf{h}}^{k\in n}$ and feature vectors $\tilde{\mathbf{h}}_{ij}$ are imposed by two $\ell_{2}$-normalization steps, as in Eq.\ref{eq:per} and Eq.~\ref{eq:across}, respectively,

\begin{equation}
{\mathbf{\hat{h}}}_{ij}=\frac{{\mathbf{h}}_{ij}}{\parallel{\mathbf{h}}_{ij}\parallel_{2}}
\label{eq:across}
\end{equation}

\begin{equation}
{\mathbf{\tilde{h}}}^{k}=\frac{\mathbf{\hat{h}}^{k}}{\parallel\mathbf{\hat{h}}^{k}\parallel_{2}}
\label{eq:per}
\end{equation}

The final normalized featuremaps $\tilde{\mathbf{h}}^{k\in n}$ are forwarded as inputs to the decoding layer of unpooling/deconvolution/nonlinearity to reconstruct the input $x$ as in Eq.~\ref{eq:decode}, 
\begin{equation}
\tilde{x}=f(\sum_{k\in n}\tilde{\mathbf{h}}^{k}\ast \mathbf{P}^{k} +c^{k})
\label{eq:decode}
\end{equation}
where $\mathbf{P}^{k}$ and $c^{k}$ are the filters and biases of decoding layer. In order to enforce the sparsity properties of (\emph{i})-(\emph{iii}), the $\ell_{1}$ sparsity is applied on $\tilde{\mathbf{h}}^{k\in n}$ as in Eq.~\ref{eq:l1-sparse}, where the averaged $\ell_{1}$ sparsity over $n$ featuremaps and $m$ training data is minimized during the reconstruction of input $x$, as in Eq{'}s.~\ref{eq:rec-loss},~\ref{eq:l1-sparse} and~\ref{eq:loss}, 

\begin{equation}
\mathcal{L}_{L2rec}=\parallel\mathbf{x}-\mathbf{\tilde{x}}\parallel_{2}
\label{eq:rec-loss}
\end{equation}

\begin{equation}
\mathcal{L}_{L1sp} = \frac{1}{m}\frac{1}{n}\sum_{d\in m}\sum_{k\in n}\parallel\mathbf{\tilde{h}}^{k}\parallel_{1}
\label{eq:l1-sparse}
\end{equation}

\begin{equation}
\mathcal{L}_{SSCAE}=\mathcal{L}_{L2rec} + \lambda_{L1sp}\mathcal{L}_{L1sp}
\label{eq:loss}
\end{equation}
where $\mathcal{L}_{L2rec}$, $\mathcal{L}_{L1sp}$ and $\mathcal{L}_{SSCAE}$ are the reconstruction, sparsity and SSCAE loss functions, respectively. $\lambda_{L1sp}$ indicates the sparsity penalty on $\tilde{\mathbf{h}}^{k\in n}$ and $\tilde{\mathbf{h}}_{ij}$. Fig.~\ref{fig:sparse} demonstrate the steps of normalization and sparsification by selected feature maps of MNIST data.

\section{Experiments}
\label{sec:experiments}

We used Theano~\cite{Bastien-Theano-2012} and Pylearn~\cite{pylearn2_arxiv_2013}, on Amazon EC2 g2.8xlarge instances with GPU GRID K520 for our experiments.

\subsection{Reducing Dead filters}
\begin{figure}[htb!]
	\centering
	\begin{minipage}{1\linewidth}
		\centering
		\includegraphics[width=12cm]{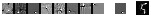}
		
		{{\footnotesize (a) CAE w/o pooling, select delta filter and featuremap}}
	\end{minipage}
	
	\begin{minipage}{1\linewidth}
		\centering
		\includegraphics[width=12cm]{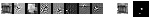}
		
		{{\footnotesize (b) SSCAE w/o pooling, select filter and sparse featuremap}}
	\end{minipage}
	
	\begin{minipage}{1\linewidth}
		\centering
		\includegraphics[width=12cm]{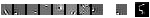}
		
		{{\footnotesize (c) CAE w/ max-pooling, select delta filter and featuremap}}
	\end{minipage}
	
	\begin{minipage}{1\linewidth}
		\centering
		\includegraphics[width=12cm]{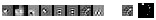}
		
		{{\footnotesize (d) SSCAE w/ max-pooling, select filter and sparse featuremap}}
	\end{minipage}
	\caption{Comparison of 8 filters learnt from MNIST by CAE and SSCAE w/o pooling (a,b) and w/ non-overlapping max-pooling (c,d) using ReLu nonlinearity. Select single filter and respective featuremaps shown on the digit.}
	\label{fig:filters}
\end{figure}

In order to compare the performance of our model in minimizing dead filters by learning sparse and local filters, the trained filters of MNIST data are compared between CAE and SSCAE with and without pooling layer in Fig.~\ref{fig:filters}. It is shown in Fig.~\ref{fig:filters}(a)(c) that CAE with and without pooling layer learn some delta filters which provide simply an identity function. However, the sparsity function used in SSCAE is trying to reduce in extracting delta filters by managing the activation across featuremaps, as shown in Fig.~\ref{fig:filters}(b)(d).  

\subsection{Improving learning of reconstruction}
\begin{figure*}[htb!]
	
	\begin{minipage}{0.45\linewidth}
		\centering
		\resizebox{1\textwidth}{!}{\includegraphics{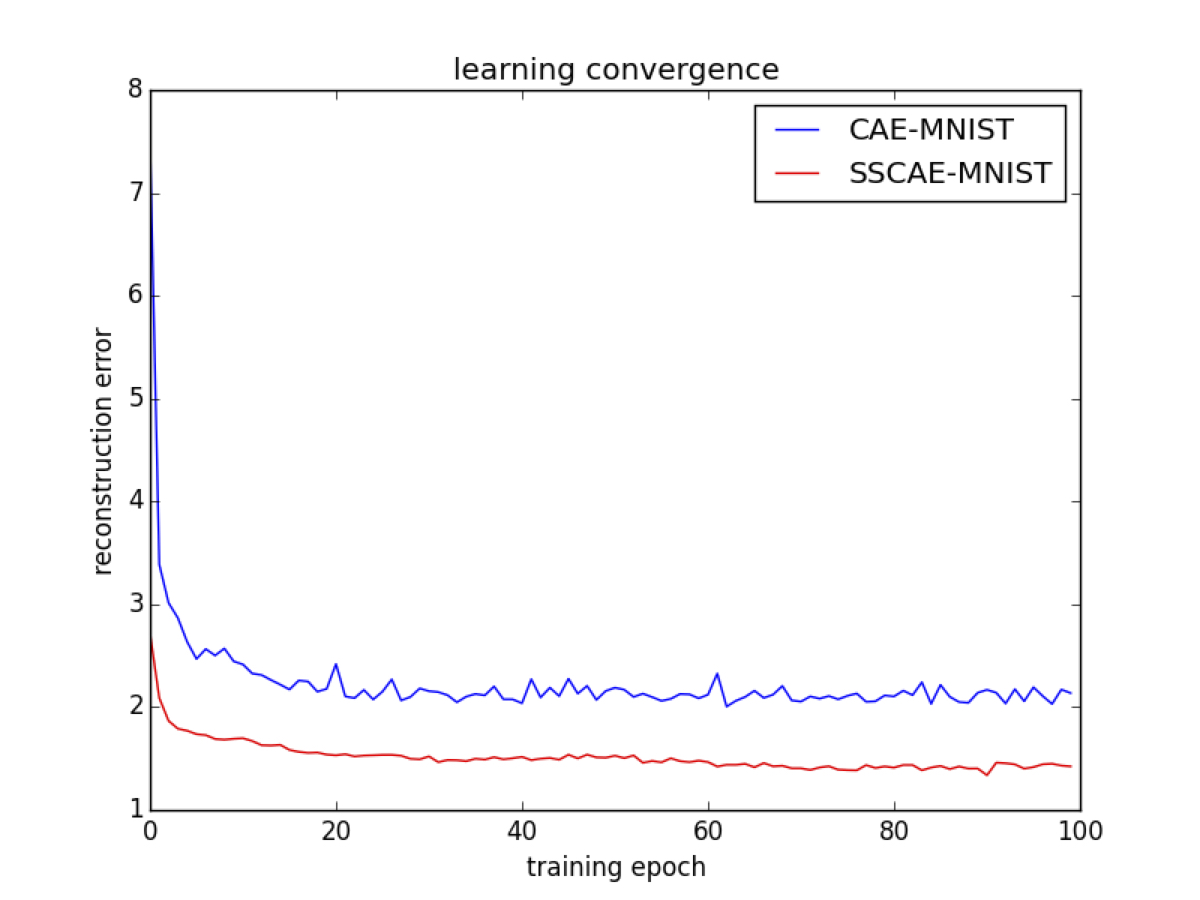}}
		
		{{\footnotesize (a)}}
	\end{minipage}
	\begin{minipage}{0.45\linewidth}
		\centering			
		\resizebox{1\textwidth}{!}{\includegraphics{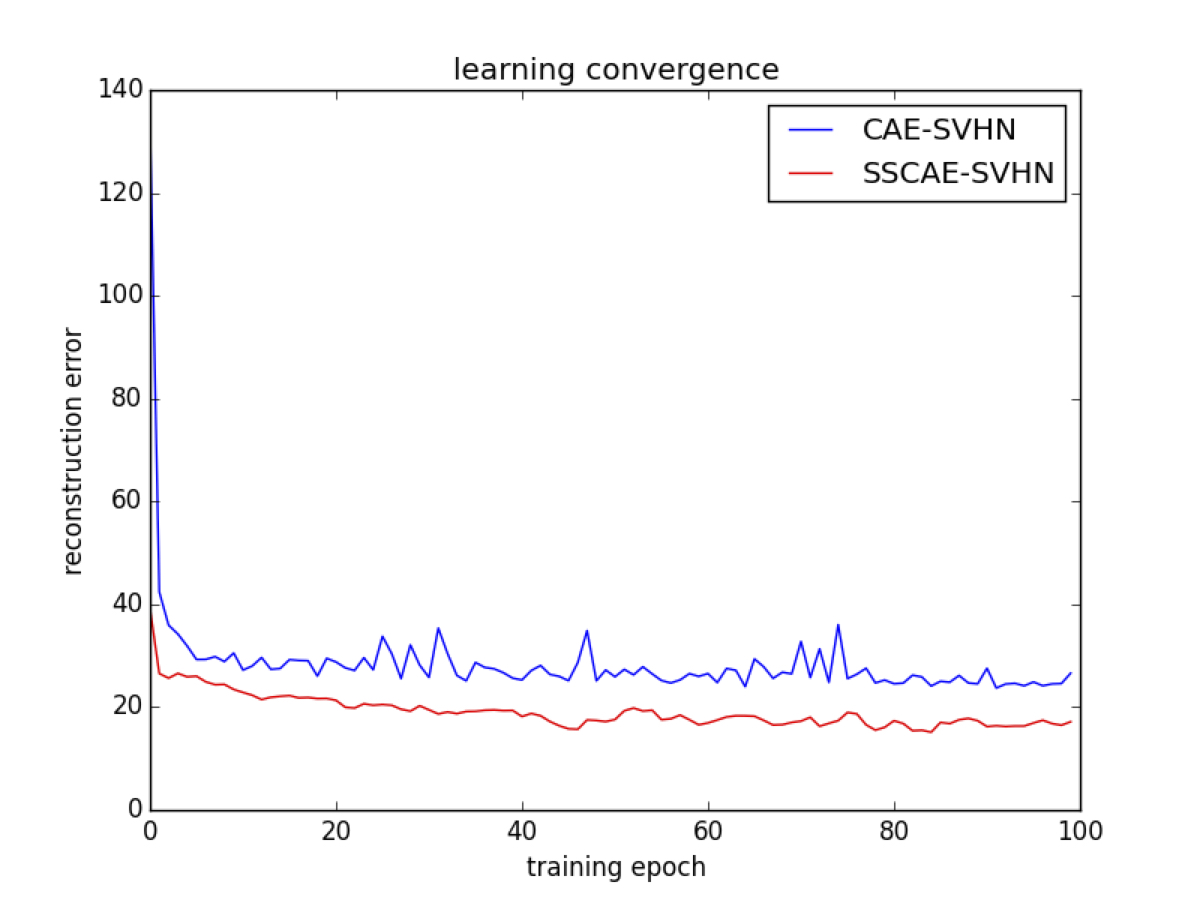}}
		
		{{\footnotesize (b)}}
	\end{minipage}
	
	\begin{minipage}{0.45\linewidth}
		\centering			
		\resizebox{1\textwidth}{!}{\includegraphics{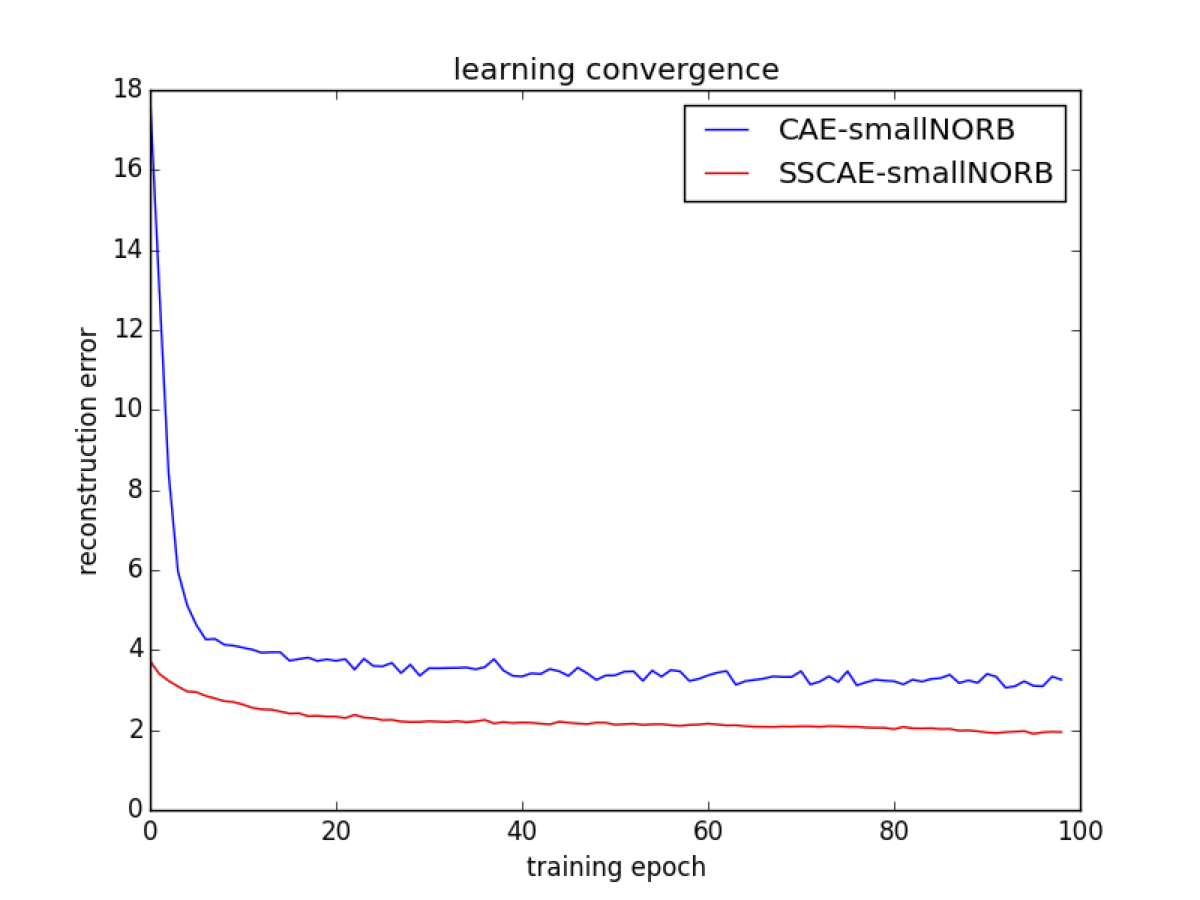}}
		
		{{\footnotesize (c)}}
	\end{minipage}
	\begin{minipage}{0.45\linewidth}
		\centering			
		\resizebox{1\textwidth}{!}{\includegraphics{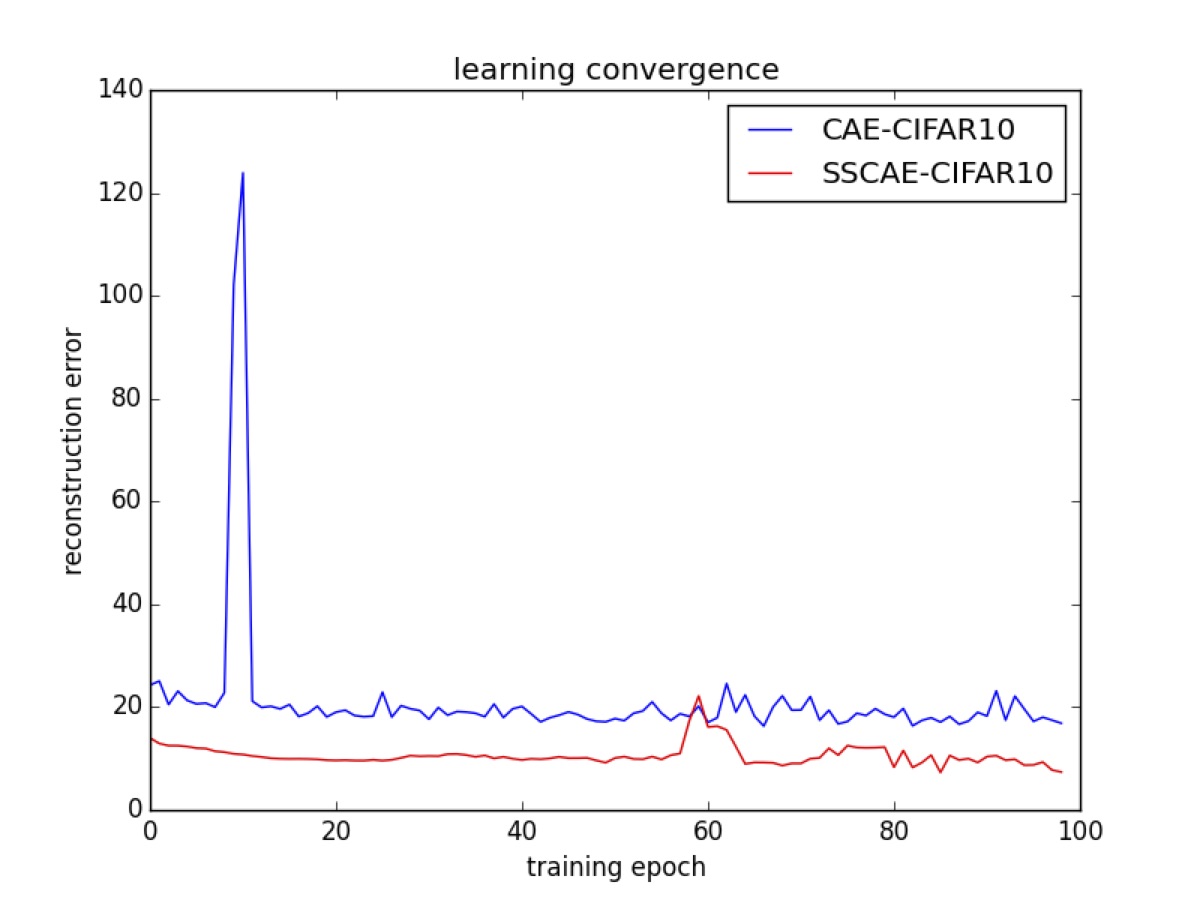}}
		
		{{\footnotesize (d)}}
	\end{minipage}
	\caption{Learning rate convergence of CAE and SSCAE on (a) MNIST, (b) SVHN, (c) small NORB, and (d) CIFAR-10 dataset using 16 filters of $11\times 11\times 3$ size.}
	\label{fig:conv-svhn}
	
\end{figure*}

\begin{figure}[htb!]
	\begin{minipage}{0.9\linewidth}
%		\vspace{-5mm}
		\centering
		\resizebox{1\textwidth}{!}{\includegraphics{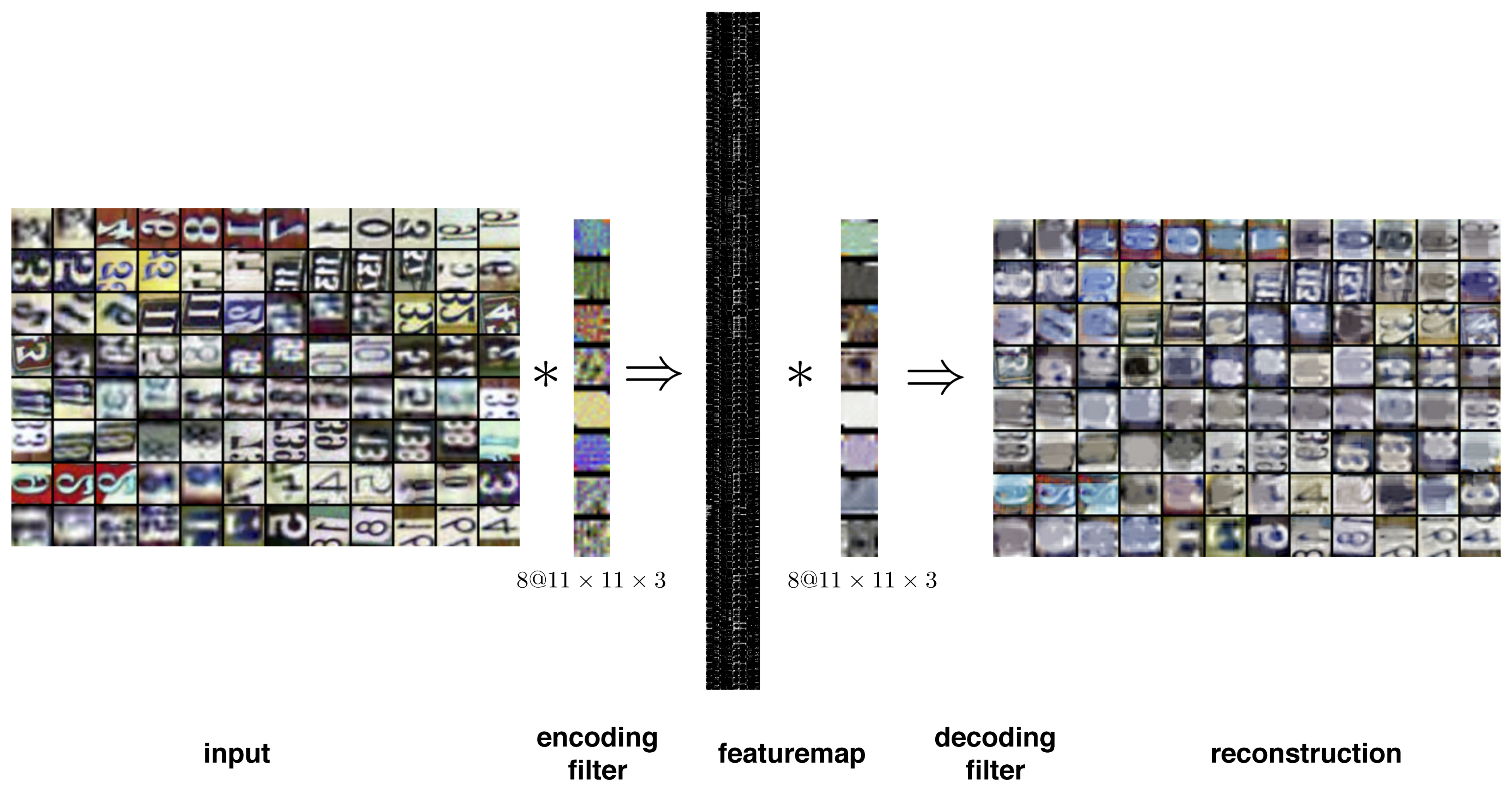}}
%		\vspace{5mm}
		{{\footnotesize (a) CAE}}
		
	\end{minipage}
	
	\begin{minipage}{0.9\linewidth}
		\centering
		\resizebox{1\textwidth}{!}{\includegraphics{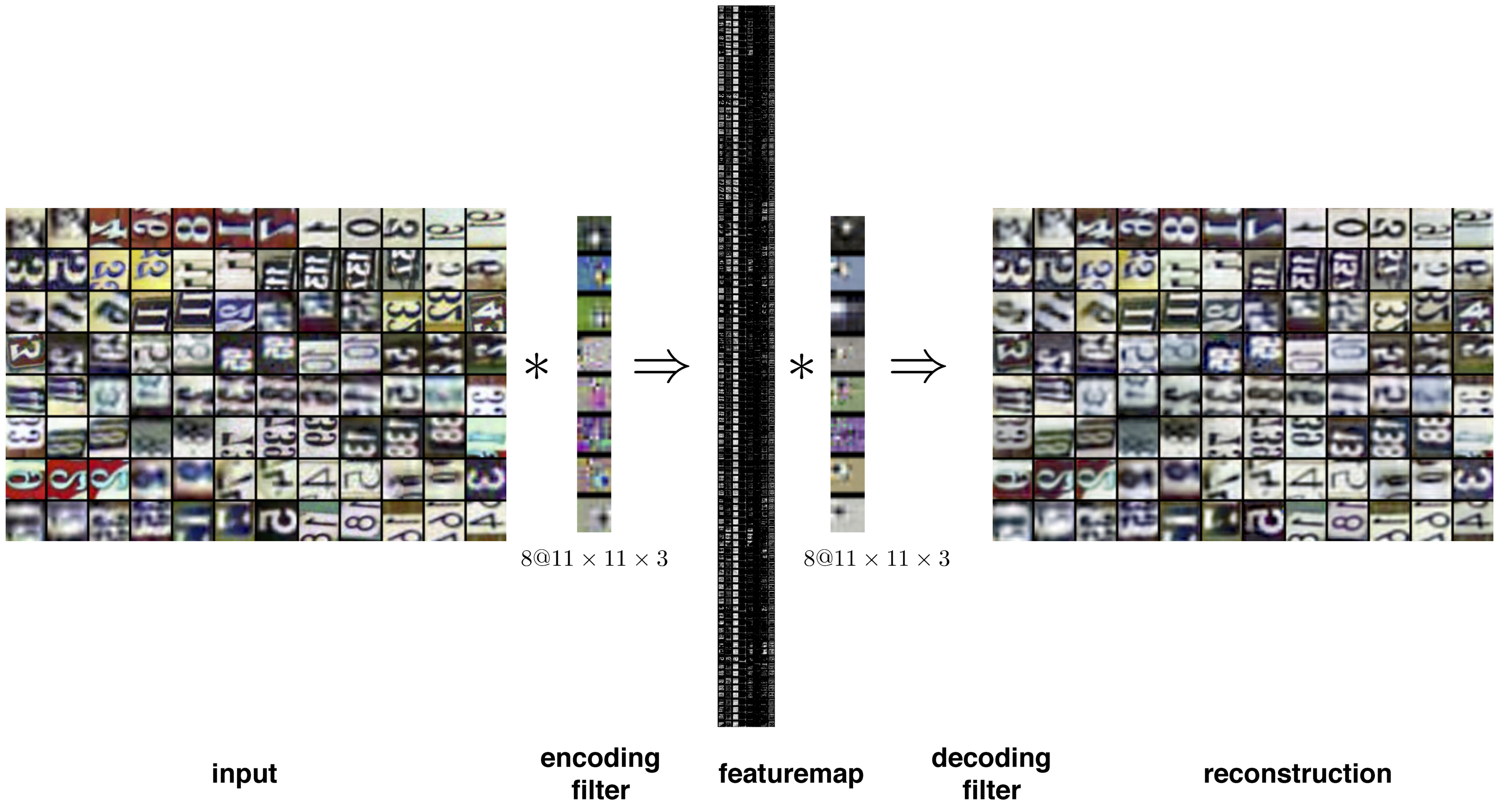}}
		
		{{\footnotesize (b) SSCAE}}
	\end{minipage}
%	\vspace{5mm}
	\caption{SVHN data-flow visualization in (a) CAE and (b) SPCAE with 8 filters. The effect of structured sparsity is shown in encoding and decoding filters and the reconstruction. NO ZCA whitening is applied.}
	\label{fig:svhn}
\end{figure}

\begin{figure}[htb!]
	\begin{minipage}{0.5\linewidth}
		%		\vspace{-5mm}
		\centering
		\includegraphics[width=5cm]{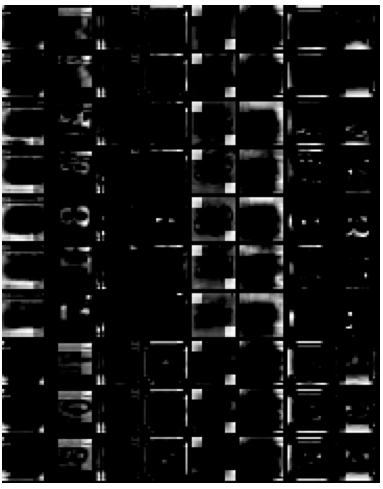}
		
		{{\footnotesize (a) CAE}}
	\end{minipage}
	\begin{minipage}{0.5\linewidth}
		\centering
		\includegraphics[width=5cm]{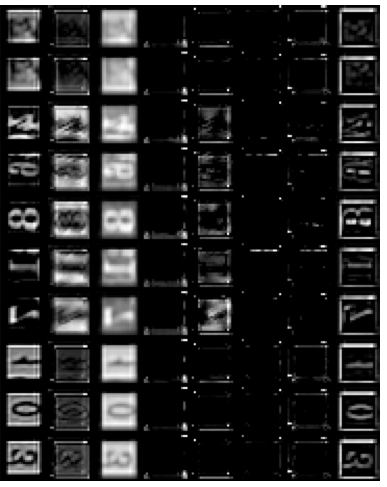}
		
		{{\footnotesize (b) SSCAE}}
	\end{minipage}
	%	\vspace{5mm}
	\caption{Selected featuremap of SVHN dataset extracted by (a) CAE, and (b)SSCAE with 8 filters of $11\times 11\times 3$ size. NO ZCA whitening is applied.}
	\label{fig:svhn_fmp}
\end{figure}

\begin{figure*}[htb!]
	
	\begin{minipage}{1\linewidth}
		\centering
		\resizebox{0.6\textwidth}{!}{\includegraphics{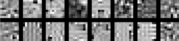}}
		
		{{\footnotesize (a) CAE encoding filter}}
	\end{minipage}
	\vspace{3mm}
	\begin{minipage}{1\linewidth}
		\centering			
		\resizebox{0.6\textwidth}{!}{\includegraphics{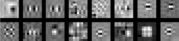}}
		
		{{\footnotesize (b) SSCAE encoding filter}}
	\end{minipage}
	\vspace{3mm}	
	\begin{minipage}{1\linewidth}
		\centering			
		\resizebox{0.6\textwidth}{!}{\includegraphics{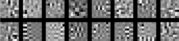}}
		
		{{\footnotesize (c) CAE decoding filter}}
	\end{minipage}
	\vspace{3mm}	
	\begin{minipage}{1\linewidth}
		\centering			
		\resizebox{0.6\textwidth}{!}{\includegraphics{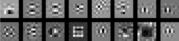}}
		
		{{\footnotesize (d) SSCAE decoding filter}}
	\end{minipage}
	\caption{16 Learnt encoding and decoding filters of (a)(c) CAE and (b)(d) SSCAE on small NORB dataset.}
	\label{fig:filter-smallnorb}
	
\end{figure*}

To investigate the effect of structured sparsity on learning of filters through reconstruction, the performance of CAE and SSCAE is compared on SVHN dataset, as shown in Fig.~\ref{fig:svhn}. To show the performance of structured sparsity on reconstruction, a small CAE with 8 filters is trained on SVHN dataset. Fig.~\ref{fig:svhn}(a) shows the performance of CAE after training which fails to extract edge-like filters and results in poor reconstruction. Fig.~\ref{fig:filter-smallnorb} also depicts the learnt 16 encoding and decoding filters on small NORB dataset, where structured sparsity improve the extraction of localized and edge-like filters. However, SSCAE outperform CAE in reconstruction due to learnt edge-like filters. The selected featuremap of the two models are shown in Fig.~\ref{fig:svhn_fmp}(a)(b). The convergence rate of reconstruction optimization for CAE and SSCAE is also compared on MNIST (Fig.~\ref{fig:conv-svhn}(a)), SVHN (Fig.~\ref{fig:conv-svhn}(b)), small NORB (Fig.~\ref{fig:conv-svhn}(c)), and CIFAR-10 (Fig.~\ref{fig:conv-svhn}(d)) datasets, which indicate faster convergence in SSCAE.

\newpage
\bibliography{references}
\bibliographystyle{iclr2016_conference}

\end{document}